\documentclass[10pt,twocolumn,letterpaper]{article}

\usepackage[pagenumbers]{cvpr}  

\definecolor{cvprblue}{rgb}{0.21,0.49,0.74}
\usepackage[pagebackref,breaklinks,colorlinks,allcolors=cvprblue]{hyperref}

\usepackage{multirow}
\usepackage{acro}
\usepackage{orcidlink}

\def\paperID{57} 
\def\confName{EarthVision}
\def\confYear{2026}

\title{Low-Data Supervised Adaptation Outperforms Prompting for Cloud Segmentation Under Domain Shift
}

\author{Harshith Kethavath\\
University of Georgia, USA\\
{\tt\small harshith@kethavath.com}
\and
Weiming Hu\\
University of Georgia, USA\\
{\tt\small weiming@uga.edu}\\
}

\DeclareAcronym{lora}{
  short = LoRA,
  long  = Low-Rank Adaptation
}

\DeclareAcronym{iou}{
    short = IoU,
    long = Intersection over Union
}

\DeclareAcronym{fft}{
    short = FFT,
    long = Full Fine-Tuning
}

\begin{document}
\maketitle

\begin{abstract}

Adapting vision-language models to remote sensing imagery presents a fundamental challenge: both the visual and linguistic distributions of satellite data lie far outside natural image pretraining corpora. Despite this, prompting remains the dominant deployment paradigm, driven by the assumption that domain-specific language can guide frozen model representations toward specialized tasks. We test this assumption directly on a domain where the mismatch is prominent: cloud segmentation for satellite imagery. Using CLIPSeg on the CloudSEN12+ cloud segmentation benchmark, we evaluate 60 prompt variants spanning simple labels, domain terminology, appearance descriptors, and contextual cues, finding that every variant underperforms the zero-shot baseline (0.255 mIoU), with engineered prompts scoring as low as 0.07 mIoU. No amount of linguistic refinement bridges the gap between CLIP's natural image representations and satellite spectral imagery. In contrast, supervised fine-tuning with just 0.1\% labeled data ($\sim$8 images) surpasses zero-shot performance overall, and 5–10\% data recovers $\sim$85\% of maximum achievable mIoU. Full fine-tuning consistently outperforms low-rank adaptation by 0.03–0.09 mIoU, with the largest gaps for spectrally ambiguous classes, and at 0.5 to 1\% labeled data, fine-tuning temporarily degrades performance on these classes before recovering, a supervision dip that aggregate mIoU can mask. For practitioners adapting vision-language models to specialized imagery, our results deliver a clear message: labeled data is not the expensive alternative to prompting; it is the worthwhile path. Our code is available at \href{https://github.com/uga-gaim/2026_cvprw_cloudprompts}{\tt https://github.com/uga-gaim/2026\_CVPRW\_C\\loudPrompts}

\end{abstract}    
\section{Introduction}
\label{sec:introduction}

Cloud detection from satellite imagery is a common task in Earth observation, yet deploying modern vision-language models for this task presents a critical assumption buried in the dominant AI deployment paradigm. A recent large-scale study found that 70\% of production AI systems rely on prompting models rather than weight tuning~\cite{pan_measuring_2025}. This preference assumes that pretrained representations are close enough to the target domain that language can bridge the remaining gap. For natural images, this assumption often holds. For satellite imagery, we show it does not.

Satellite observations differ from natural photographs in fundamental ways. Visually, overhead perspectives, multi-spectral sensors, and amorphous atmospheric phenomena, like clouds that blend into haze, shadows without hard edges, bear little resemblance to the object-centric natural scenes that dominate vision-language pretraining. Linguistically, the gap is equally severe; meteorological vocabulary like ``optically thin cirrus'' or ``cloud shadow'' rarely appears in the image caption pairs used to train CLIP-based models. This dual shift creates a compound mismatch that prompting alone cannot resolve.

We investigate this failure mode using CLIPSeg~\cite{luddecke_image_2022}, a promptable segmentation model trained on PhraseCut~\cite{wu_phrasecut_nodate}, a dataset of natural images annotated with phrases like ``the brown dog'' or ``leftmost chair'', and evaluate pretrained and fine-tuned variants of the model on CloudSEN12+~\cite{aybar_cloudsen12_2024}, the largest expert-labeled cloud segmentation dataset for Sentinel-2 satellite imagery. Our central question: \textbf{when severe domain shift is present, can prompt engineering alone compensate for it, or is supervised fine-tuning necessary considering various levels of annotation cost?}

We conduct a controlled empirical comparison across 60 prompt variants, \ac{lora}, and \ac{fft} across data budgets from 0.1\% to 100\%. The answer is unambiguous: every engineered prompt underperforms simple label baselines, while supervised fine-tuning with just 8 labeled images surpasses zero-shot performance on average. Small labeled datasets are not a last resort; they are the right first choice when severe domain shift is present. Our contributions are therefore threefold:
\begin{enumerate}
    \item We establish that linguistic refinement cannot compensate for fundamental visual-linguistic domain shift, providing the first systematic evidence of this failure for satellite segmentation.
    \item We identify a surprisingly low supervision crossover point, as few as 0.1\% labeled data ($\sim$8 images) suffice to outperform any prompt strategy, making the case for zero-shot deployment difficult to justify. We further identify a supervision dip phenomenon, at 0.5–1\% labeled data, fine-tuning temporarily degrades performance on spectrally ambiguous classes (thin cloud, cloud shadow) before recovering at 2.5–5\%, revealing that aggregate mIoU can mask class-level harm when annotation budgets are extremely tight.
    \item We show that the choice between \ac{lora} and \ac{fft} is not a compute tradeoff but a task structure decision: spectral ambiguity, not data volume, determines where each method succeeds.
\end{enumerate}
\section{Related Work}
\label{sec:relatedwork}

\subsection{Prompt Engineering}

Prompt engineering for vision-language models has been studied extensively in image classification, where template-based prompting consistently outperforms raw label prompts~\cite{radford_learning_2021}. Learnable prompt methods such as CoOp and CoCoOp extend this by optimizing prompt tokens end-to-end, achieving strong generalization across classification benchmarks~\cite{zhou_conditional_2022, khattak_maple_2023}. DenseCLIP has begun extending prompt conditioning to dense prediction~\cite{rao_denseclip_2022}, though systematic evaluation for segmentation remains limited.

The existing literature leaves two areas under-investigated. First, prompt learning evaluations nearly always operate within natural image domains, testing generalization to novel classes or related datasets but not to fundamentally different visual modalities. Second, no prior work systematically evaluates prompt engineering for segmentation under the dual visual-linguistic shift of satellite imagery, where overhead perspectives, spectral sensors, and meteorological vocabulary all diverge from pretraining distributions simultaneously. We fill both gaps, and find that failure under this shift is total and consistent.

\subsection{Supervised Adaptation}

Supervised adaptation of models spans a spectrum from \ac{fft}, which updates all parameters for maximum flexibility at the cost of compute and forgetting risk~\cite{khazem_topolora-sam_2026}, to parameter-efficient methods like \ac{lora}, which inject low-rank updates into transformer layers while keeping backbone weights frozen~\cite{hu_lora_2021}. The performance tradeoffs between these approaches vary by domain and data regime~\cite{shuttleworth_lora_2025}.

Empirically, \ac{lora} has demonstrated competitive performance with \ac{fft} on dense prediction tasks including segmentation, with recent work showing it can match unconstrained optimization when adaptation targets are within the low-rank subspace's expressive range~\cite{khazem_topolora-sam_2026}. \ac{lora}'s adoption has been further driven by minimal inference overhead; updates merge directly into pretrained weights, and a compact hyperparameter space that facilitates systematic search~\cite{belanec_peft-bench_2025}.

What remains unclear is how these adaptation strategies compare across varying annotation budgets (availability of labelled data) for vision-language segmentation in remote sensing. This is a domain where visual and linguistic distributions diverge substantially from pretraining data. Prior work does not establish where the performance crossover from zero-shot to supervised adaptation occurs, nor how much labeled data each method requires before gains saturate. Our experiments directly address both questions.

\subsection{Domain Shift in Remote Sensing}

Vision-language models learn joint representations by aligning images and text during pretraining, but this alignment is distribution specific. CLIP's contrastive objective trains on natural image-caption pairs which reflect the associations between visual patterns and linguistic concepts. When target domains diverge substantially from pretraining distributions, the learned alignment may not transfer, even if underlying visual representations remain useful. This has motivated domain-specific variants such as RemoteCLIP, RS-CLIP, and SenCLIP~\cite{li_rs-clip_2023, liu_remoteclip_2024, jain_senclip_nodate}. However, such adaptation often requires substantial data curation and computing resources that might be unavailable to practitioners developing the models.

A complementary question remains unanswered: for practitioners without access to domain-adapted pretraining, what is the most effective adaptation strategy given a fixed annotation budget? Prior work does not characterize whether prompt engineering can bridge the gap for existing models, nor at what data threshold supervised adaptation becomes worthwhile. Our work addresses this directly, establishing both the failure ceiling of prompting and the minimum supervision needed to surpass it, for the widely deployed CLIP-based segmentation family.
\section{Methodology}
\label{sec:methodology}

\subsection{Dataset}

We evaluate on CloudSEN12+~\cite{aybar_cloudsen12_2024}, a large-scale dataset for cloud and cloud shadow detection in Sentinel-2 imagery. It is the largest expert labeled dataset for this task, containing image patches distributed globally across all continents except Antarctica. Each patch is 509x509 pixels at a 10-meter resolution, captured from Sentinel-2.

The dataset provides pixel-wise semantic labels for four classes: clear sky, thick cloud, thin cloud, and cloud shadow. This four-class schema captures the diversity of atmospheric phenomena that challenge remote sensing applications. We use the high quality annotation subset, which contains expert-reviewed pixel-level labels. Using the MLSTAC format from Hugging Face, we obtain 8,490 training patches, 535 validation patches, and 975 test patches. For our experiments, we use only RGB bands (B4, B3, B2) to maintain compatibility with CLIPSeg, which expects three channel input. All reported metrics are computed on the held out test set.

\subsection{Model}

CLIPSeg~\cite{luddecke_image_2022} represents a foundational architectural pattern that underlies a broad family of vision-language segmentation models, including LSeg, OpenSeg, SegCLIP, MaskCLIP, and OVSeg~\cite{li_language-driven_2022, avidan_scaling_2022, luo_segclip_nodate, avidan_extract_2022, liang_open-vocabulary_2023}. All share a common design, the frozen CLIP encoder whose representations are conditioned on text prompts, paired with a lightweight task-specific decoder. Critically, all inherit the same vulnerability: their backbones are pretrained exclusively on natural image-caption pairs, with no exposure to satellite imagery, spectral data, or meteorological concepts. Findings on CLIPSeg therefore speak to this entire architectural family, not a single model.

We use the \texttt{clipseg-rd64-refined} variant, which pairs a CLIP ViT-B/16 visual encoder with a lightweight transformer decoder of dimension 64, processing images at 352×352 resolution. The architecture keeps the CLIP backbone frozen while training only a compact decoder ($\sim$1.1M parameters), making it well suited for isolating adaptation strategy effects. CLIPSeg was trained on PhraseCut~\cite{wu_phrasecut_nodate}, a natural image dataset annotated with common English phrases with no references to satellite imagery, no spectral observations, and no meteorological terminology.

This complete absence of remote sensing exposure is precisely what makes CLIPSeg an ideal testbed. Unlike domain-adapted variants such as RemoteCLIP or RS-CLIP~\cite{liu_remoteclip_2024, li_rs-clip_2023}, CLIPSeg allows us to measure the domain gap in its unmitigated form, establishing a clear lower bound for prompting and a clean baseline from which supervised adaptation gains can be accurately attributed. Domain-adapted variants require large-scale remote sensing corpora and substantial compute for pretraining, resources unavailable to most practitioners, and thus fall outside the deployment scenario this work targets.

To establish precise terminology used throughout our experiments: the zero-shot baseline refers to the pretrained CLIPSeg checkpoint used without any additional training, where class predictions are obtained by applying argmax over logits from four class-specific text prompts at inference. \ac{fft} updates all model parameters, allowing unconstrained representational adaptation. \ac{lora} is applied exclusively to the decoder's attention projection matrices, keeping the backbone frozen, making it a more parameter-efficient alternative.

\subsection{Low-Rank Adaptation}

To compare parameter-efficient adaptation with \ac{fft}, we use \ac{lora} ~\cite{hu_lora_2021}, a method that freezes pretrained weights and injects trainable low-rank decomposition matrices into transformer layers. For a pretrained weight matrix $W_0 \in \mathbb{R}^{d\times k}$, \ac{lora} parameterizes the update as $W_0 + \Delta W = W_0 + BA$, where $B \in \mathbb{R}^{d\times r}$ and $A \in \mathbb{R}^{r\times k}$ with rank $r \ll min(d,k)$. This reduces trainable parameters by orders of magnitude while preserving the pretrained model's knowledge. In essence, rather than updating the full weight matrix directly, \ac{lora} aims to learn a low-rank (memory-efficient) approximation of these updates with a neural network, capturing the necessary task-specific adaptations with far fewer parameters.

Recent work has extended \ac{lora} to dense prediction tasks including semantic segmentation, where it successfully adapts foundation models to specialized domains such as medical imaging and remote sensing~\cite{zhong_convolution_2024}. We apply \ac{lora} to query, key, value and output projection matrices ($W_q$, $W_k$, $W_v$, $W_o$) of CLIPSeg's transformer decoder, following standard practice for attention-based adaptation. Hyperparameter selection is detailed in Section 3.4.

\subsection{Hyperparameter Configuration}

We conduct hyperparameter searches for both \ac{fft} and \ac{lora} to identify optimal configurations before the low-data sweep experiments.

\begin{table}[h]
    \centering
    \begingroup
    \renewcommand{\arraystretch}{1.5}
    \resizebox{\columnwidth}{!}{%
        \begin{tabular}{cccccc}
          \hline
          \textbf{\ac{fft}} & Learning Rate ($\times 10^{-5}$) & $1$ & $2$ & $5$ & $10$ \\
          \hline
          \multirow{2}{*}{\textbf{LoRA}} & Learning Rate ($\times 10^{-5}$) & $5$ & $10$ & $20$ \\
          \cline{2-5}
                                  & Rank & $8$ & $16$ & $32$ \\
          \hline
        \end{tabular}
    }
    \endgroup
    \caption{Variables in Hyperparameter Search in FFT and LoRA}
    \label{hyper_parameter_search}
\end{table}

For \ac{fft}, we explore learning rates while fixing other hyperparameters as 20 epochs, weight decay of 0.02, and a warm-up ratio of 0.06. Table~\ref{hyper_parameter_search} reports the values of learning rates, with $5 \times 10^{-5}$ achieving the highest validation mIoU on the validation set. For \ac{lora}, we search over learning rates and ranks, setting $\alpha = 2r$ throughout. We fix \ac{lora} dropout at 0.05, weight decay at 0.01, warm-up ratio at 0.03, and train for 15 epochs. Table~\ref{hyper_parameter_search} presents the values of learning rates and ranks in the grid search, with learning rate $2 \times 10^{-4}$ and rank $32$ yielding the best validation performance. These optimal configurations are frozen for all subsequent low-data experiments (Section 4.2), where we vary only the percentage of training and validation data to isolate the effect of annotation cost on performance.

\subsection{Loss Function}

We train CLIPSeg using a composite loss function designed for binary segmentation under class imbalance:
\begin{equation*}
\resizebox{\columnwidth}{!}{$
  L = w_{\text{focal}} \cdot L_{\text{focal}} + w_{\text{tversky}} \cdot L_{\text{tversky}} + w_{\text{boundary}} \cdot L_{\text{boundary}}
$}
\end{equation*}

Focal loss~\cite{lin_focal_2018} addresses the severe foreground background imbalance, which is inherent in one vs rest cloud segmentation by down-weighting well classified pixels, with parameters $\alpha = 0.75$ and $\gamma = 2.0$. Tversky loss~\cite{salehi_tversky_2017} generalizes Dice loss with asymmetric weighting of false positives and false negatives ($\alpha_T = 0.3$, $\beta_T = 0.7$), penalizing missed detections more heavily, critical for thin clouds and shadows which occupy small image regions compared to clear and thick cloud. Boundary loss applies morphological edge detection to upweight pixels near class boundaries by a factor of $2$, improving delineation of cloud edges.

The component weights ($w_{focal}$ = 0.8, $w_{tversky}$ = 1.0, $w_{boundary}$ = 0.1) were fixed across all experiments. The dominant Tversky weight reflects our priority on minimizing missed detections for minority classes, while focal loss receives substantial weight to ensure stable training when most pixels are easily classified. Boundary weighting is kept low to improve edge sharpness without over segmenting clouds and shadows, which naturally lack well-defined edges. This loss function is identical for both \ac{lora} and \ac{fft}, ensuring fair comparison between adaptation strategies.

\subsection{Evaluation Metrics}

We evaluate segmentation performance using \ac{iou}, computed per-class as:
\[IoU_c = \frac{TP_c}{TP_c + FP_c + FN_c}\]
where $TP_c$, $FP_c$, and $FN_c$ denote true positives, false positives, and false negatives for class $c$. Mean IoU (mIoU) is the unweighted average across all four classes. We also report per-class IoU to analyze performance on spectrally challenging categories such as thin cloud and cloud shadow.

Because CLIPSeg only supports binary segmentation, at inference, we obtain class predictions by applying argmax over the logits from all four class-specific prompts, producing a single multiclass segmentation mask. All metrics are computed on the held out test set (975 images) by accumulating a global confusion matrix across all samples.
\section{Experiments}
\label{sec:experiments}

\subsection{Prompt Sensitivity Analysis}

Prompt engineering fails for CLIPSeg on satellite imagery, consistently and without exception across every strategy we evaluated. We designed 15 prompt variants per class spanning simple labels, domain terminology, appearance descriptors, and contextual phrases, producing 60 total combinations. Every variant underperforms the zero-shot baseline of simple class labels (0.255 mIoU), with some prompts scoring as low as 0.07 mIoU, a 73\% relative degradation.

\begin{table}[h]
    \centering
    \begingroup
    \renewcommand{\arraystretch}{1.5}
    \resizebox{\columnwidth}{!}{%
        \begin{tabular}{clcccc}
        \hline
        & \multicolumn{4}{c}{Prompts} & \\
        \cmidrule(lr){2-5}
        Variant (mIoU) & Clear & Thick Cloud & Thin Cloud & Cloud Shadow \\
        \hline
        V1 (0.222)  & ground        & white cloud        & thin cloud              & shadow  \\
        V2 (0.196) & earth         & bright white cloud & wispy cloud             & dark shadow \\
        V7 (0.068) & ground below  & cloud              & faded cloud             & dark patch  \\
        V8 (0.216) & visible ground & overhead cloud    & soft cloud              & sharp shadow  \\
        V10 (0.108) & open ground   & thick cloud        & light cloud             & ground shadow  \\
        V11 (0.196) & clear ground & white dense cloud & cloud veil & cloud shadow \\
        \hline
        \end{tabular}
    }
    \endgroup
    \caption{Representative prompt variants and mIoU. All variants 
    underperform the zero-shot baseline (0.255 mIoU). Baseline 
    prompts: clear, thick cloud, thin cloud, cloud shadow.}
    \label{prompt_variants}
\end{table}

Table~\ref{prompt_variants} presents a representative subset of the 60 variants tested. Prompts were designed to span four strategies: minimal single-word labels, domain-specific terminology, appearance descriptors, and contextual phrases. For spectrally distinct classes, variants ranged from single words (``cloud'', ``white'') to multi-word descriptions (``bright white opaque cloud''). For ambiguous classes, variants targeted transparency (``wispy cloud'', ``semi-transparent cloud''), spatial relationships (``shadow beneath cloud'', ``ground shadow''), and surface identity (``terrain'', ``landscape''). Cumulative prompts that progressively added context also consistently underperformed across all classes.

The most revealing failure comes from exclusionary prompts. Negative formulations such as ``not cloud'' and ``not haze'' produced the worst results across all variants. This failure is architecturally grounded. Although CLIP's contrastive training includes negative examples, which are mismatched image-text pairs within each batch, but these teach the model that certain images and captions are unrelated, not how to interpret the word ``not'' as a semantic operator~\cite{radford_learning_2021}. The text encoder was never exposed to captions like ``not cloud'' paired with cloud-free images. The token ``not'' carries no learned visual meaning; the embedding remains dominated by ``cloud.'' This is not a prompt design failure; however, it is a fundamental property of how CLIP's embedding space was constructed. Learnable prompt methods such as CoOp optimize within this same embedding space and therefore face the same representational ceiling: the bottleneck is the visual encoder's misalignment with satellite spectral imagery, not the prompt strategy.

\begin{figure}[h]
  \centering
   \includegraphics[width=\linewidth]{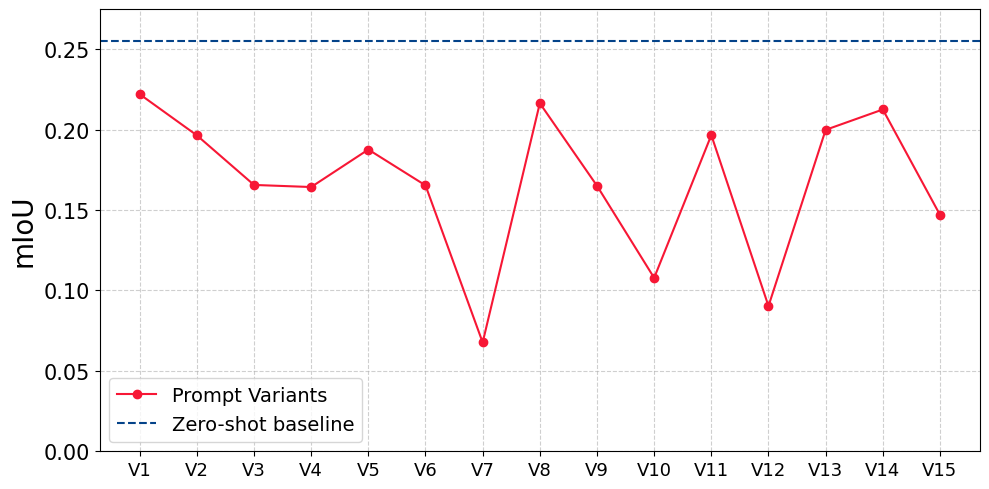}
   \caption{Mean IoU across 15 prompt variant combinations. The dashed line indicates the zero-shot baseline (0.255 mIoU). Every engineered variant falls below the baseline.}
   \label{fig:prompt_variants}
\end{figure}

Figure~\ref{fig:prompt_variants} presents mIoU across all 15 variant combinations. Every engineered variant underperforms the simple label baseline, with no prompt strategy recovering meaningful performance. These results establish that CLIPSeg's representations are too misaligned with satellite imagery for language alone to bridge the gap.

\subsection{Annotation Efficiency}

Having established that prompt engineering alone cannot bridge the domain gap, we turn to the complementary question: how much labeled data is required to do so? The answer is remarkably little. Both \ac{lora} and \ac{fft} surpass the zero-shot baseline with as few as 0.1\% of the training set, approximately 8 images. This finding fundamentally challenges the assumption that annotation cost justifies zero-shot deployment, as the crossover point from prompting to supervised adaptation requires negligible labeling effort.

To ensure robustness at low data regimes where subset composition can dominate results, we perform 10 independent runs at each data percentage using different random seeds for subset sampling, reporting averaged metrics across runs. We evaluate data budgets from 0.1\% to 100\% of the training set (approximately 8 to 8,490 images), spanning the full range from minimal to complete supervision.

\begin{figure}[h]
  \centering
   \includegraphics[width=\linewidth]{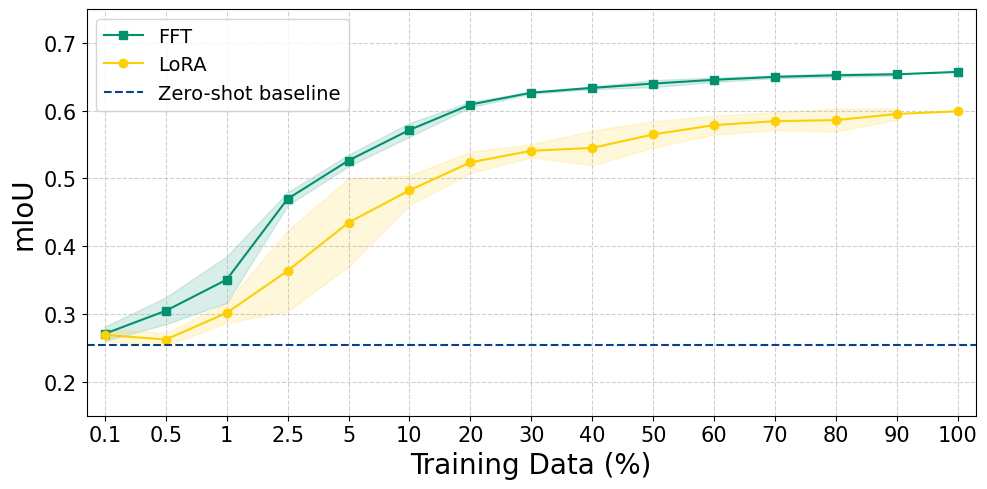}
   \caption{Mean IoU as a function of training data percentage for LoRA and FFT. The dashed line indicates the zero-shot baseline (0.255 mIoU). Shaded regions indicate standard deviation across 10 independent runs.}
   \label{fig:low_data_curves_shaded_std}
\end{figure}

Figure~\ref{fig:low_data_curves_shaded_std} shows mIoU as a function of training data for both methods. Performance follows a logarithmic growth pattern, that is rapid gains occur in the low-data regime, with diminishing returns beyond 30\% data. \ac{fft} reaches 0.57 mIoU at 10\% data and 0.66 mIoU at 100\%; \ac{lora} achieves 0.48 and 0.60 mIoU at the same checkpoints. The gap between methods remains stable at 0.04–0.07 mIoU throughout, a consistency we return to in Section 4.3, suggesting that \ac{fft}'s advantage stems from representational capacity rather than differential data efficiency.

\begin{figure}[h]
  \centering
   \includegraphics[width=\linewidth]{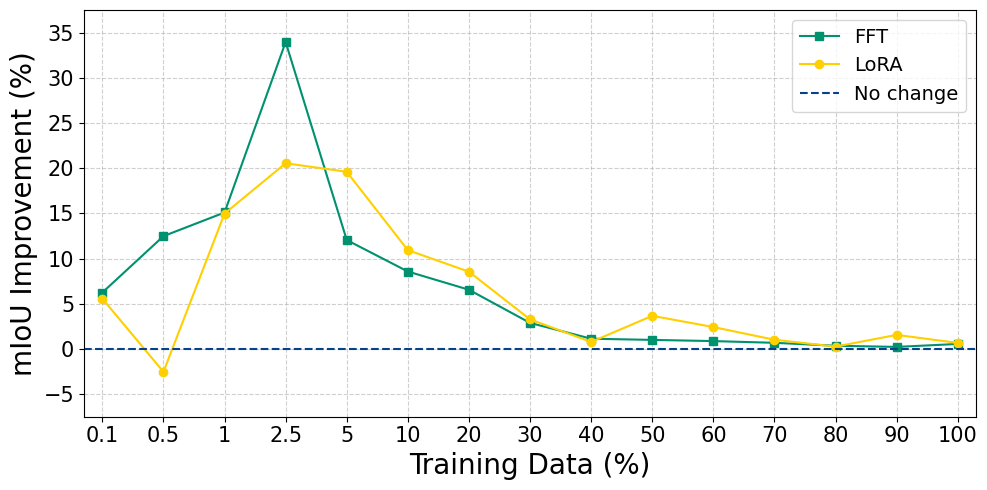}
   \caption{Marginal mIoU improvement at each data increment relative to the previous checkpoint. Peak efficiency occurs at 2.5\% data. LoRA exhibits instability at 0.5\% (negative improvement).}
   \label{fig:low_data_improvement_curves}
\end{figure}

\begin{figure*}[h]
  \centering
   \includegraphics[width=\linewidth]{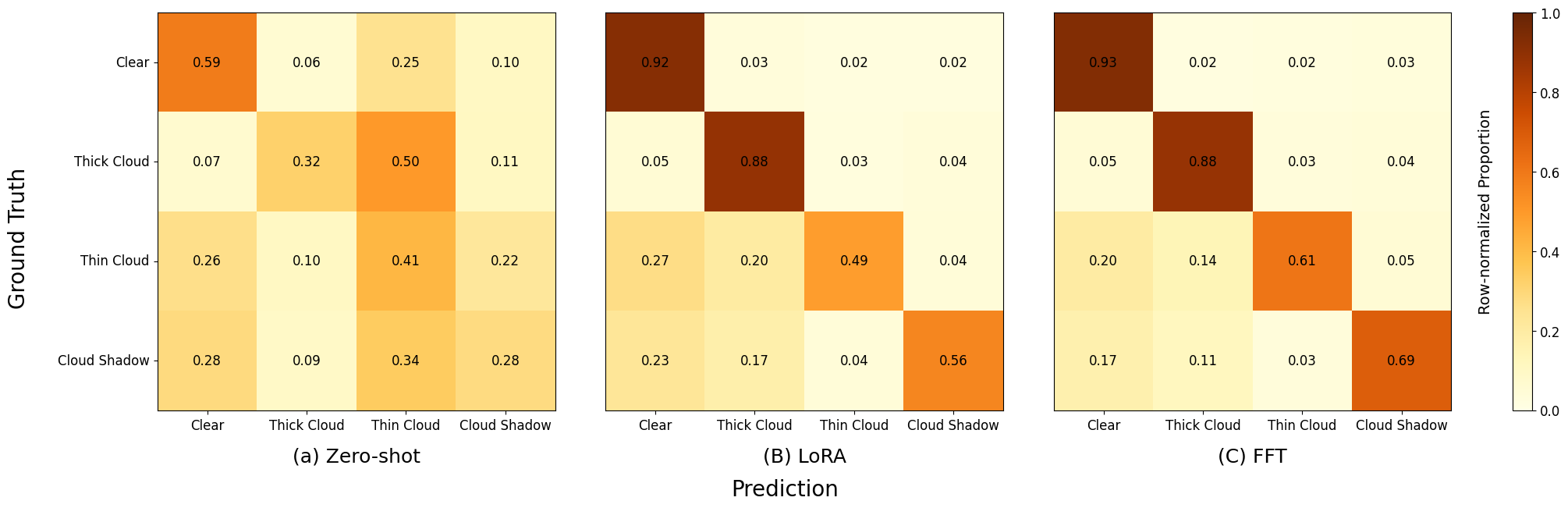}
   \caption{Row-normalized confusion matrices for (a) zero-shot, (b) LoRA, and (c) FFT at 100\% training data. Diagonal entries represent per-class correct classification rates.}
   \label{fig:confusion_matrices}
\end{figure*}

Figure~\ref{fig:low_data_improvement_curves} presents marginal mIoU improvement at each data increment relative to the previous checkpoint. Both methods peak in annotation efficiency at 2.5\% data, \ac{fft} achieves 34\% improvement over the prior checkpoint, \ac{lora} achieves 21\%. Beyond this point, each additional labeled sample yields progressively smaller returns, with marginal gains falling below 3\% after 30\% data. One important asymmetry emerges: \ac{lora} exhibits negative improvement at 0.5\% data, an instability absent in \ac{fft}. This reflects a minimum supervision threshold specific to parameter-efficient adaptation, with too few examples, \ac{lora}'s low-rank updates are pulled toward noise rather than signal, while \ac{fft}'s unconstrained parameter space remains stable. Practitioners using \ac{lora} should treat 1\% labeled data as a practical minimum.

For remote sensing practitioners facing annotation constraints, these results provide a clear target: labeling 5–10\% ($\sim$425 to 850 images) of available data recovers approximately 85\% of maximum achievable mIoU while requiring a fraction of the full annotation effort. Beyond 30\% ($\sim$2,500 images), additional labels yield marginal returns that rarely justify the annotation cost.

\subsection{\ac{fft} vs Low-Rank Adaptation}

The consistent 0.03–0.09 mIoU gap between \ac{fft} and \ac{lora} observed in Figure~\ref{fig:low_data_curves_shaded_std} raises a more specific question than simple performance ranking: is this gap uniform across classes, or concentrated where the task is hardest? The confusion matrices in Figure~\ref{fig:confusion_matrices} suggest, the gap is driven by two spectrally ambiguous classes: thin cloud and cloud shadow.

Zero-shot predictions establish the baseline difficulty. Cloud shadow is distributed nearly uniformly across all classes (recall: 0.28, 0.09, 0.34, 0.28), and thick cloud is misclassified as thin cloud 50\% of the time, confirming the domain mismatch established in Section 4.1. Both fine-tuning methods resolve these confusions substantially.

For spectrally distinct classes, \ac{lora} and \ac{fft} perform nearly identically. Clear sky correct classification rate rises from 0.59 to 0.92 (\ac{lora}) and 0.93 (FFT); thick cloud from 0.32 to 0.88 for both methods. The gap between adaptation strategies is negligible, parameter efficiency is sufficient when the classification boundary is visually unambiguous. For spectrally ambiguous classes the picture changes sharply. Thin cloud correct classification rate reaches 0.49 under \ac{lora} but 0.61 under \ac{fft}, a 12 point gap. Cloud shadow shows a 13 point gap (0.56 vs. 0.69). These are not small differences for classes that directly affect downstream Earth observation applications.

This divergence reflects a fundamental constraint of low-rank adaptation. \ac{lora} restricts weight updates to a low-rank subspace, which is sufficient when the target classification boundary is spectrally distinct: clear sky and thick cloud occupy separable regions of the embedding space that low-rank updates can reach. Thin cloud and cloud shadow require fine-grained reshaping of representations to capture subtle spectral overlap between semi-transparent cloud layers and underlying surface reflectance, and between shadow regions and dark terrain. These distinctions are multi-dimensional in embedding space and exceed what a constrained low-rank subspace can express. \ac{fft}, unconstrained, reshapes freely and captures them.

For practitioners, the implication is specific: when targeting well-defined classes, \ac{lora}'s computational efficiency makes it the practical choice. When thin cloud or cloud shadow detection is the priority, as in most atmospheric correction and Earth observation pipelines, \ac{fft}'s additional parameter cost is justified by the performance gap. Section 4.4 examines how these class-level differences evolve across data regimes.

\subsection{Per-Class Analysis}

\begin{figure*}[h]
  \centering
   \includegraphics[width=\linewidth]{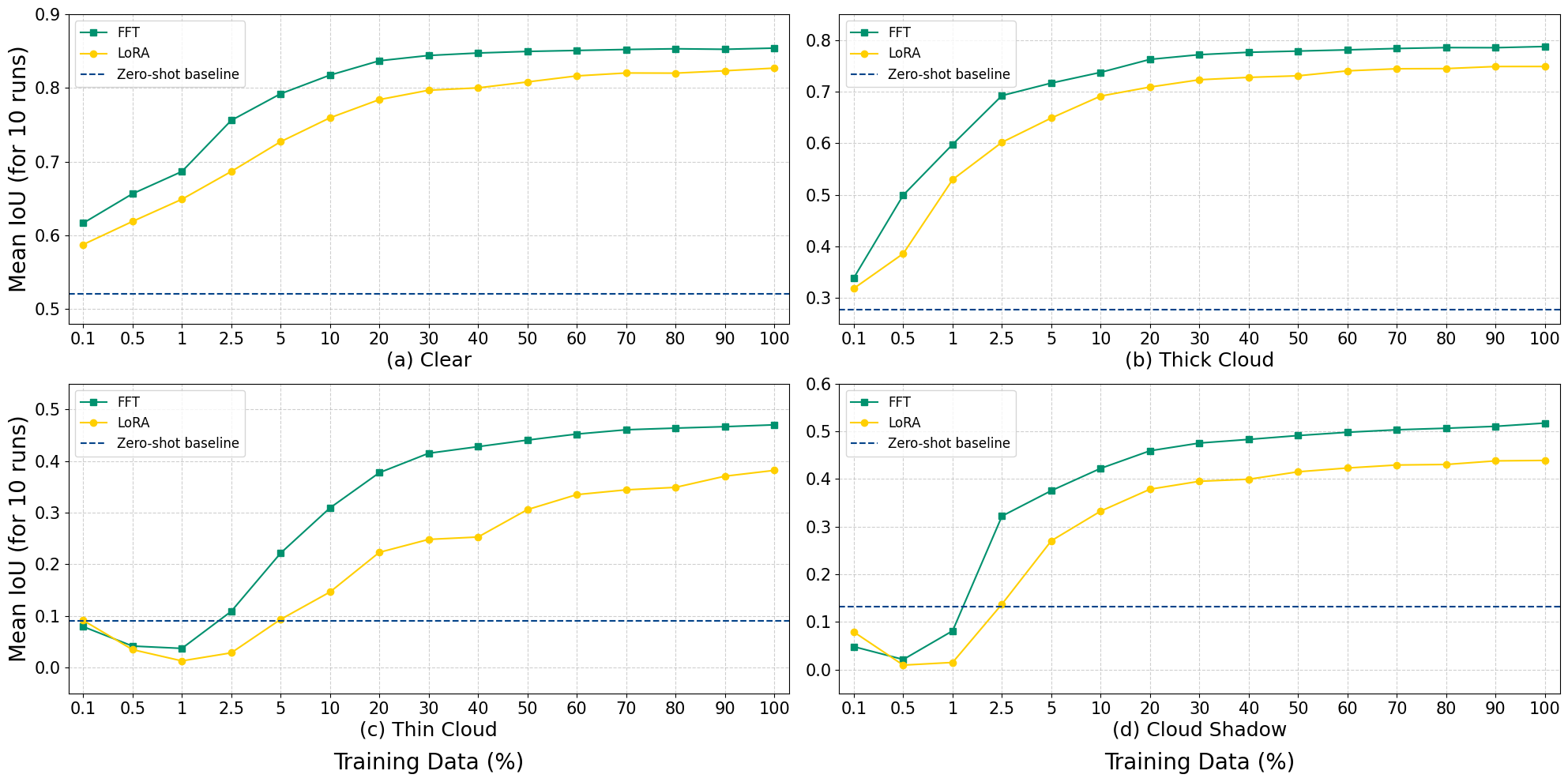}
   \caption{Per-class IoU as a function of training data percentage for LoRA and FFT. Dashed lines indicate per-class zero-shot baselines.}
   \label{fig:per_class_low_data_curves}
\end{figure*}

The four atmospheric classes divide into two distinct regimes that persist across all adaptation strategies and data budgets. Clear sky and thick cloud, which are spectrally distinct with visually unambiguous signatures, respond immediately to supervision and reach high performance with minimal data. Thin cloud and cloud shadow, which are spectrally overlapping with adjacent classes and absent from CLIP's pretraining distribution, exhibit fundamentally different learning patterns that Figure~\ref{fig:per_class_low_data_curves} makes visible. At 100\% training data, Figure~\ref{fig:per_class_low_data_curves} confirms this two-regime structure: clear sky and thick cloud reach 0.83–0.85 and 0.75–0.79 mIoU respectively, consistent with the confusion matrix results in Section 4.3, while thin cloud and cloud shadow remain lower at 0.38–0.47 and 0.44–0.52 mIoU, confirming that spectral ambiguity imposes a performance ceiling that persists even at full data.

Figure~\ref{fig:per_class_low_data_curves} also reveals a more nuanced finding invisible in aggregate mIoU, that for thin cloud and cloud shadow, supervised adaptation initially degrades performance before improving it. Both classes drop below their zero-shot baselines at 0.5–1\% labeled data before recovering at 2.5–5\%. This temporary degradation reflects a specific failure mode of low-data fine-tuning for ambiguous classes: with too few representative examples, the supervision signal is insufficient to reshape embeddings toward the target distribution, but strong enough to disrupt the zero-shot embedding structure that provided whatever minimal signal existed at baseline. The result is the worst of both regimes, the model loses zero-shot coherence without gaining supervised accuracy. Practically, this suggests that practitioners fine-tuning with fewer than 1\% labeled samples should monitor per-class performance rather than aggregate mIoU, as overall gains may mask temporary degradation on minority classes.

\begin{figure*}[h]
  \centering
   \includegraphics[width=\linewidth]{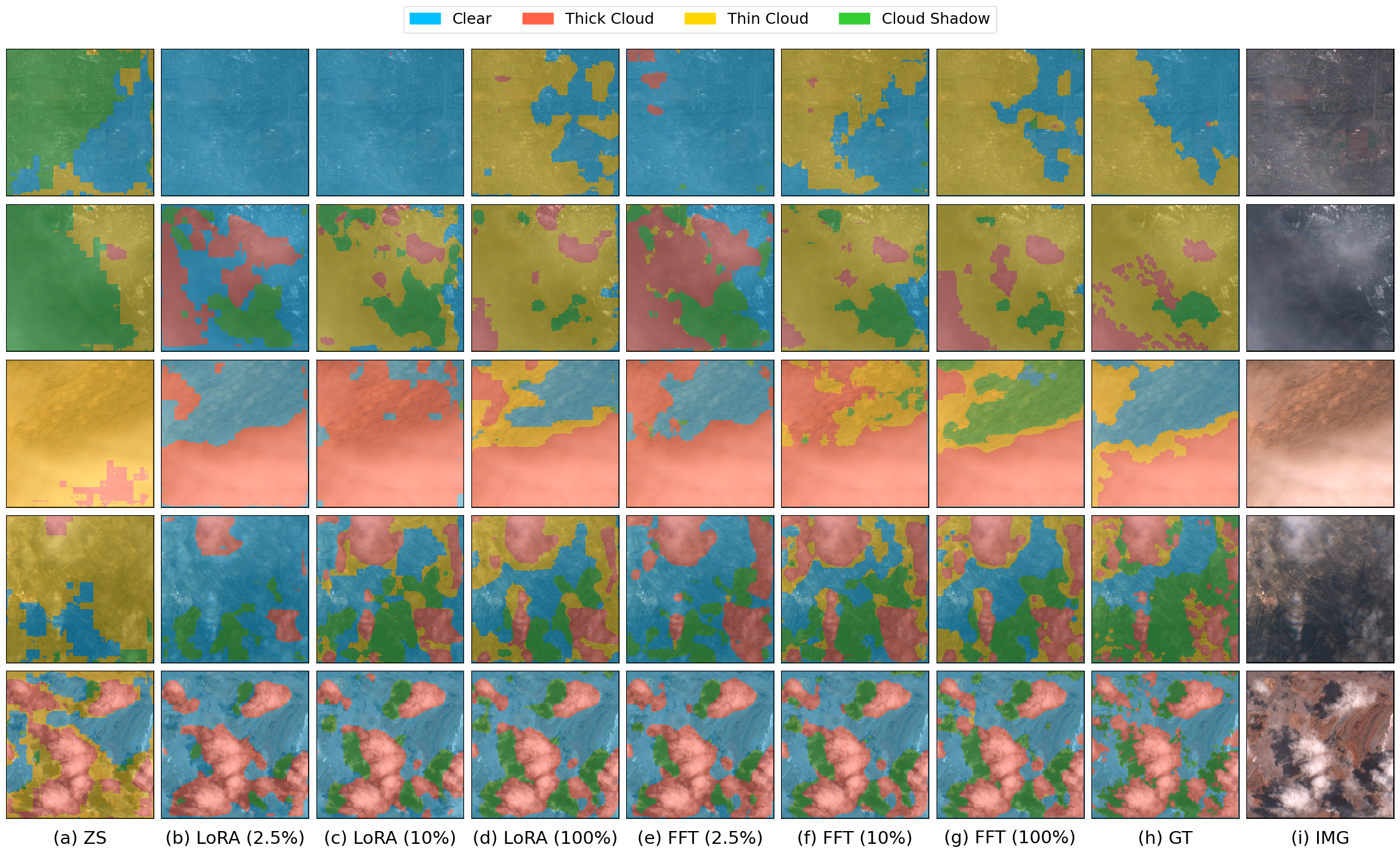}
   \caption{Qualitative segmentation results across five test samples. Columns show zero-shot (ZS), LoRA at 2.5\%/10\%/100\%, FFT at 2.5\%/10\%/100\%, ground truth (GT), and input RGB image (IMG).}
   \label{fig:samples}
\end{figure*}

Figure~\ref{fig:samples} illustrates these dynamics qualitatively across five test samples. Zero-shot predictions exhibit systematic class confusion, cloud shadow predicted over clear sky regions, thin cloud bleeding into thick cloud boundaries, consistent with the near-zero IoU baselines. At 2.5\% data, coarse class structure is recovered by both methods, though thin cloud boundaries and cloud shadow delineation remain imprecise, particularly for \ac{lora}. At 100\% data, \ac{fft} closely matches ground truth across all samples; \ac{lora} produces comparable results but with residual under-segmentation of cloud shadow at boundaries.

\subsection{Limitations and Future Work}

\textbf{Architectural Scope.} We focus on CLIPSeg as a deliberate representative of the frozen-CLIP-encoder family of segmentation models, a design shared by others (Section 3.2). Our generalizability argument rests on the shared vulnerability of CLIP backbones pretrained on natural images. Empirical validation on architectures with meaningfully different decoder mechanisms, would strengthen this claim and constitutes a direct next step.

\noindent
\textbf{Dataset and Sensor Scope.} All experiments use CloudSEN12+, a globally distributed expert-labeled dataset for Sentinel-2 imagery. The global distribution likely makes our low-data subsets more representative than locally collected alternatives, which may mean the 2.5–5\% efficiency threshold is optimistic for datasets with regional bias or lower annotation quality. Whether this threshold transfers to other remote sensing domains depends on task structure. While cross-sensor validation remains as future work, CLIPSeg's architectural representativeness make the current scope a principled starting point. Extending to multispectral input incorporating SWIR and NIR bands, which carry discriminative signal for spectrally ambiguous classes, represents a natural next step but would require architectural modification to CLIPSeg's three-channel input.

\noindent
\textbf{Class Imbalance Effects.} CloudSEN12+ exhibits natural imbalance, clear sky and thick cloud dominate over thin cloud and cloud shadow. Our composite loss function addresses imbalance during training, but not during low-data subset sampling: at 0.5–1\% data, minority class examples may be too sparse to provide stable supervision, directly contributing to the supervision dip observed in Section 4.4. Stratified sampling strategies that guarantee minority class representation in low-data regimes represent a straightforward mitigation.

\noindent
\textbf{Manual Prompt Engineering Scope.} Our evaluation covers 60 variants spanning all major manual prompt design strategies. Learnable prompt tuning methods like CoOp, CoCoOp, and prompt ensembling, represent a distinct paradigm we do not evaluate. However, as argued in Section 4.1, these methods optimize within the same misaligned embedding space, and empirical verification of this prediction remains valuable future work.

\noindent
\textbf{Future Directions.} Three extensions follow directly from our findings. First, developing adaptation strategies that address \ac{lora}'s performance gap on spectrally ambiguous classes, informed by our analysis of where low-rank constraints fail, could yield parameter-efficient methods competitive with \ac{fft} for atmospheric segmentation. Second, semi-supervised approaches leveraging abundant unlabeled satellite imagery alongside minimal labels may reduce annotation requirements further and could mitigate the dip phenomenon by providing richer distributional coverage at very low data regimes. Third, extending to multi-temporal cloud detection, where temporal consistency provides additional supervisory signal, could improve performance on cloud shadow specifically.
\section{Conclusion}
\label{sec:conclusion}

We set out to test a foundational assumption of modern AI deployment, that pretrained vision-language models can be guided to specialized domains through careful prompting. For satellite imagery cloud segmentation, this assumption fails completely. The domain gap between CLIP's natural image pretraining and Sentinel-2 spectral imagery is not a gap that language can bridge.

Three findings define what does work. First, every one of 60 engineered prompt variants underperforms simple class-label baselines, with the worst scoring 0.07 mIoU against a 0.255 baseline, which is a 73\% relative degradation that is total and consistent across all prompt strategies. Second, supervised fine-tuning surpasses zero-shot performance with just 8 labeled images, and 5–10\% ($\sim$ 425 to 850 images) of the training set recovers approximately 85\% of maximum achievable mIoU, a remarkably low annotation threshold. Third, the choice between \ac{lora} and \ac{fft} is not a compute tradeoff but a task structure decision: for spectrally distinct classes both methods perform equivalently, but for thin cloud and cloud shadow, the spectral overlap demands representational reshaping exceeding what low-rank constraints can express, and hence requires \ac{fft}'s unconstrained parameter space.

For the Earth observation community, these results carry a specific message. Cloud detection underpins atmospheric correction across virtually every downstream EO application. A few hundred expert-labeled patches, modest by any annotation standard, can lead to performance that no prompt engineering strategy can approach. In specialized imagery domains, labeled data might not continue to be the expensive alternative to prompting. It is the worthwhile path.

{
    \small
    \bibliographystyle{unsrtnat}
    \bibliography{main}
}

\end{document}